\documentclass{llncs} %

\usepackage{float}

\usepackage{booktabs}
\usepackage{graphicx}
\usepackage{wrapfig}
\usepackage{comment}
\usepackage[table]{xcolor}

\usepackage[labelfont=bf]{caption}
\usepackage[colorlinks=true,linkcolor=blue,citecolor=blue]{hyperref}

\usepackage{color,soul}
\usepackage{amsmath} 
\usepackage{mathtools} 
\usepackage{amsfonts} 
\usepackage{amssymb}

\usepackage{array}
\usepackage{booktabs}
\usepackage{rotating}
\usepackage{multirow}
\usepackage{multicol}
\usepackage{lscape}
\usepackage{subfig}

\usepackage[export]{adjustbox}

\usepackage[ruled,vlined]{algorithm2e}

\definecolor{violet}{rgb}{0.54, 0.17, 0.89}
\definecolor{coral}{rgb}{1.0, 0.25, 0.25}
	
\usepackage{xargs}                      

\usepackage[colorinlistoftodos,prependcaption,textsize=tiny]{todonotes}
\usepackage[colorlinks=true,linkcolor=blue]{hyperref}%

\usepackage{soul}

\usepackage{tikz,xcolor,hyperref}

\definecolor{lime}{HTML}{A6CE39}
\DeclareRobustCommand{\orcidicon}{
	\begin{tikzpicture}
	\draw[lime, fill=lime] (0,0) 
	circle [radius=0.16] 
	node[white] {{\fontfamily{qag}\selectfont \tiny ID}};
	\draw[white, fill=white] (-0.0625,0.095) 
	circle [radius=0.007];
	\end{tikzpicture}
	\hspace{-2mm}
}

\foreach \x in {A, ..., Z}{\expandafter\xdef\csname orcid\x\endcsname{\noexpand\href{https://orcid.org/\csname orcidauthor\x\endcsname}
			{\noexpand\orcidicon}}
}
			
\usepackage{color}
\definecolor{reddish}{rgb}{0.82, 0.1, 0.26}

\begin{document}

\title{Recurrent Multigraph Integrator Network for Predicting the Evolution of Population-Driven Brain Connectivity Templates} 

\titlerunning{Short Title}  

\author{Oytun Demirbilek\index{Demirbilek, Oytun}  \and Islem Rekik\orcidA{} \index{Rekik, Islem}\thanks{ {corresponding author: irekik@itu.edu.tr, \url{http://basira-lab.com}. }}}
  
\institute{BASIRA Lab, Faculty of Computer and Informatics, Istanbul Technical University, Istanbul, Turkey}

\maketitle              

\begin{abstract} 
Learning how to estimate a connectional brain template (CBT) from a population of  brain multigraphs, where each graph (e.g., functional) quantifies a particular relationship between pairs of brain regions of interest (ROIs), allows to pin down the unique connectivity patterns shared across individuals. Specifically, a CBT is viewed as an integral representation of a set of highly heterogeneous graphs and ideally meeting the centeredness (i.e., minimum distance to all graphs in the population) and discriminativeness (i.e., distinguishes the healthy from the disordered population) criteria. So far, existing works have been limited to only integrating and fusing a population of brain multigraphs acquired at a single timepoint. In this paper, we unprecedentedly tackle the question: \emph{``Given a baseline multigraph population, can we learn how to integrate and forecast its CBT representations at follow-up timepoints?''} Addressing such question is of paramount in predicting common alternations across healthy and disordered populations. To fill this gap, we propose Recurrent Multigraph Integrator Network (ReMI-Net), the \emph{first graph recurrent neural network} which infers the baseline CBT of an input population $t_1$ and predicts its longitudinal evolution over time ($t_i > t_1$). Our ReMI-Net is composed of recurrent neural blocks with graph convolutional layers using a cross-node message passing to first learn hidden-states embeddings of each CBT node (i.e., brain region of interest) and then predict its evolution at the consecutive timepoint. Moreover, we design a novel time-dependent loss to regularize the CBT evolution trajectory over time and further introduce a cyclic recursion and learnable normalization layer to generate well-centered CBTs from time-dependent hidden-state embeddings. Finally, we derive the CBT adjacency matrix from the learned hidden state graph representation. ReMI-Net significantly outperformed benchmark methods in both centeredness and discriminative connectional biomarker discovery criteria in demented patients. Our ReMI-Net GitHub code is available at \url{https://github.com/basiralab/ReMI-Net}.

\end{abstract} 
 
\keywords{Longitudinal multigraphs  $\cdot$ Recurrent graph convolution $\cdot$ Graph population template evolution forecasting $\cdot$ Connectional brain template}

\section{Introduction}

The development of network neuroscience \cite{Bassett:2017} aims to present a holistic representation of the brain graph (also called network or connectome), a universal map of heterogeneous pairwise brain region relationships (e.g., correlation in neural activity or dissimilarity in morphology). Due to its multi-fold complexity, the underlying causes of neurological and psychiatric disorders, such as Alzheimer's disease, autism, and depression remain largely unknown and difficult to pin down \cite{Fornito:2015,Heuvel:2019}. How these brain disorders unfold at the individual and population scales remains one of the most challenging obstacles to understanding how the \emph{brain graph} gets altered by disorders, let alone a \emph{brain multigraph}. Conventionally, a brain multigraph is composed of a set of graphs, each capturing a unique `view' of the brain wiring network (such as morphology or function) \cite{Bassett:2017,Mheich:2020}. A single view of the brain graph is represented as a symmetric adjacency matrix where each element stores the connectivity weight between a pair of anatomical regions of interest (ROIs), encapsulating a particular type of interaction between them. Learning how to integrate a population of brain multigraphs remains a formidable challenge to identify the most representative and shared brain alterations caused by a specific disorder. Very recent works addressed this challenge by learning a \emph{connectional brain template} --in short CBT-- from a heterogeneous population of brain multigraphs. Such integral and compact encoding of a connectomic population of brain graphs into a single connectivity matrix (i.e., CBT) presents a powerful and easy tool for comparing connectomic populations of brain multigraphs \cite{HCP,CRHD,ADdataset} in different states (e.g., healthy and disordered) \cite{Dhifallah:2020,Gurbuz:2020}. A well-representative CBT is well-centered (i.e., minimum distance to all graphs in the population) and discriminative (i.e., distinguishes the healthy from the disordered population) \cite{Dhifallah:2020}.

So far, existing CBT learning methods \cite{Dhifallah:2020,Mhiri:2020,Gurbuz:2020}  have been limited to only integrating and fusing a population of brain multigraphs acquired at a single timepoint, limiting their generalizability to \emph{longitudinal} (i.e., time-series) multigraph populations. Ideally, one would design a model to not only integrate a multigraph population at a single baseline timepoint $t_1$, but also \emph{forecast} the time-dependent evolution of the learned baseline CBT at follow-up timepoints $t_i > t_1$. Adding to the difficulty of integrating a set of heterogeneous brain connectomes, predicting the future of a population via the compact CBT representation, to eventually map out and discover disorder-specific alternations, presents a big jump in the field of network neuroscience, which we set out to take in this paper. \textbf{To the best of our knowledge, no existing works attempted to solve the challenging problem of CBT evolution prediction from a baseline multigraph population observed at $t_1$.} Recently, \cite{Dhifallah:2020} proposed the netNorm framework that leverages similarity network fusion \cite{Wang:2014} for integrating multi-view brain graphs by building a high-order population graph capturing the most centered brain connectivity weights across individuals. Although pioneering, such method resorted to using Euclidean distance for graph comparison which overlooks the non-Euclidean nature of a graph as well as its node-specific topological properties. To address this limitation, recently \cite{Mhiri:2020} proposed a supervised multi-topology network cross-diffusion (SM-Net Fusion) to learn a population CBT using a weighted combination of the multi-topological matrices that encapsulate the various topologies in a heterogeneous graph population. Nonetheless, SM-Net Fusion can only handle a population of graphs derived from a single view or neuroimaging modality, failing to generalize to multigraphs.  To better model the complex interactions at the \emph{individual} brain multigraph level (i.e., between ROIs) and at the \emph{population} level (i.e., between multigraphs), one can leverage the power of the emerging graph neural networks (GNNs) \cite{GNN2018review,GNN2019review,Bessadok:2021} in learning end-to-end mapping for our CBT integration and evolution forecasting tasks. Very recently, \cite{Gurbuz:2020} proposed deep graph normalizer (DGN), the state-of-the-art method to integrate a population of brain multigraphs into a representative CBT using GNNs. Although compelling, DGN remains a fully integrational --and not predictive-- GNN architecture, that is not customized to \emph{dynamic} multigraph populations.

To address all these limitations, we propose our \textbf{Recurrent Multi-graph Integrator Network (ReMI-Net)}, the first recurrent graph convolutional neural network architecture for \emph{integrating} a baseline multigraph population into a CBT and \emph{predicting} its time-dependent evolution in a progressive manner. \emph{First}, our ReMI-Net inputs a baseline population of multigraphs and learns how to integrate them into a baseline CBT representation which evolves with time through our recurrent graph convolutional block. Specifically, our model includes a learnable view-normalization block since each view of the individual multigraph can differ in scale and distribution from other views. This learns how to integrate the multigraph population into a baseline CBT at initial timepoint $t_1$. \emph{Second}, we propose a novel \emph{graph-based} recurrent block that predicts the baseline CBT evolution by learning a low-dimensional hidden-state vector for each node (ROI) at each timepoint. Each time-specific cell of our recurrent block is a message passing network that captures the non-linearities between brain regions (i.e., multigraph nodes) in a graph convolutional end-to-end fashion. \emph{Third}, for each subject in the population and at each timepoint $t_i \geq t_1$, we pass the hidden state embeddings of each node (ROI) to next cell to learn its embedding at the consecutive timepoint $t_{i+1}$. \emph{Fourth}, we design a novel time-dependent loss to regularize the CBT evolution trajectory over time. \emph{Finally}, we derive the CBT connectivity matrix at timepoint $t_i$ by computing the pairwise distance between the hidden-state node embeddings. We present several major contributions to the state-of-the-art at three different levels. \emph{(1) Methodological level.} Our ReMI-Net is the first GNN architecture for brain CBT learning and time-dependent trajectory prediction from a baseline timepoint. We design a novel message passing network with recurrent multigraph inputs and derive our CBT from the learned time-dependent hidden-state node embeddings. We also propose a novel time-dependent loss to regularize the forecasted CBTs over time. \emph{(2) Conceptual level.} Our framework introduced the concept of a longitudinal network template (i.e., `atlas') evolution prediction from a baseline multigraph population. \emph{(3) Clinical level.} We demonstrate that ReMI-Net generates reliable and biologically sound brain templates that fingerprint the temporal evolution of demented brains and reveal time-dependent biomarkers.

\begin{figure}
\centering
\includegraphics[width=12cm]{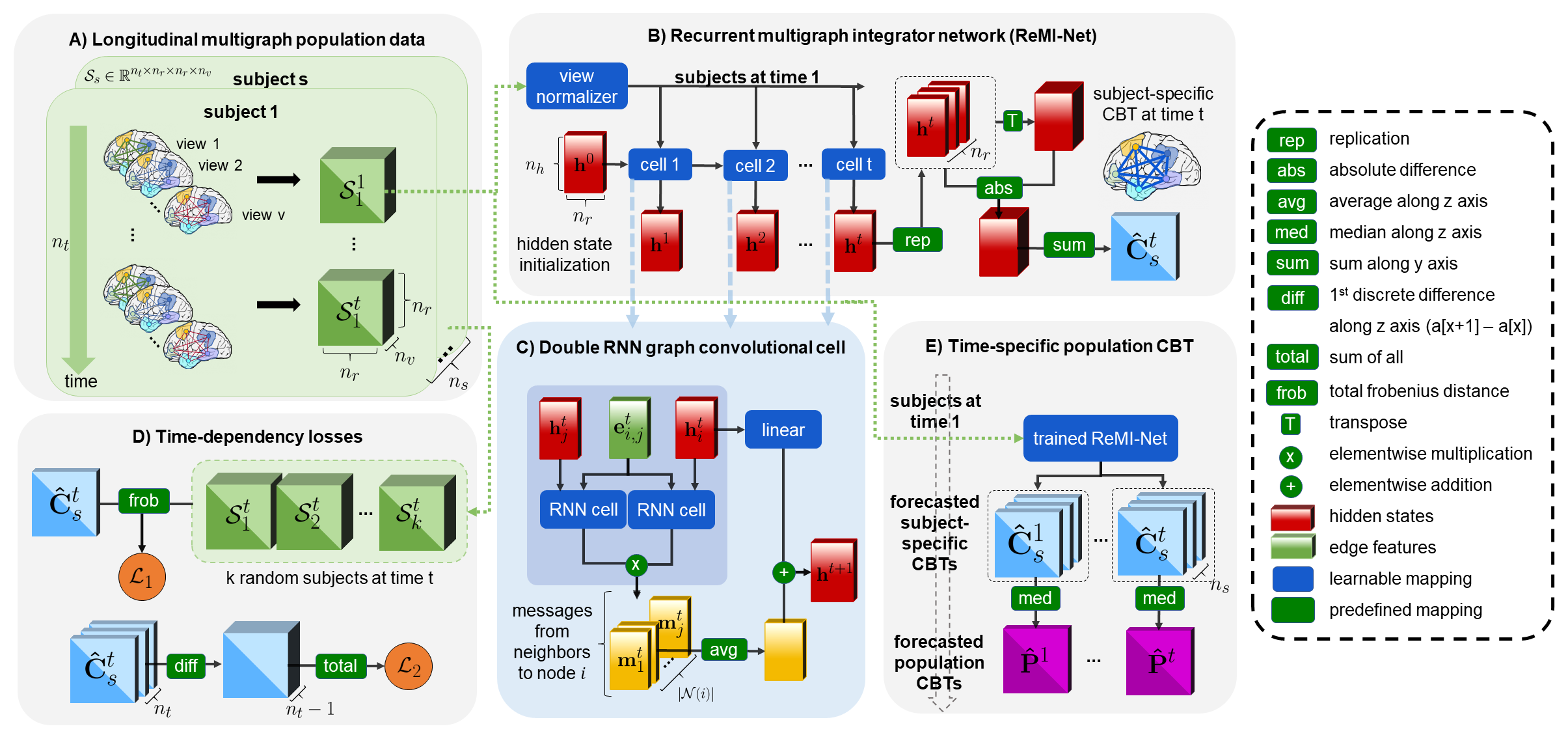}
\caption{\emph{Overview of our ReMI-Net architecture for integrating a population of multigraphs into a representative CBT and forecasting its evolution over time.} Each block of our architecture is detailed in the Method section with reference to components in this figure.} 
\label{fig1}
\end{figure}

\section{Proposed Method}

In the following, we explain the main blocks of the proposed ReMI-Net architecture\footnote{\url{https://github.com/basiralab/ReMI-Net}} for forecasting representative, centered and discriminative CBTs from a given multigraph population at baseline timepoint. \textbf{Fig.}~\ref{fig1} displays the five key blocks of our  ReMI-Net: \textbf{A)} Tensor representation of longitudinal multigraph data, \textbf{B)} Prediction of the \emph{subject-specific} CBT from learned recurrent hidden-state node embeddings, \textbf{C)} Novel message passing network to perform recurrent convolutional operations on multigraphs and generate hidden-state node embeddings, \textbf{D)} Proposed CBT centeredness loss $\mathcal{L}_c$ and time regularization loss $\mathcal{L}_t$, and \textbf{E)} Generation of the \emph{population-specific} CBT evolution trajectory. 
 
\textbf{A) Longitudinal multigraph population data.} Let $\mathbb{G}_c = \{ \mathcal{S}^t_1, \dots, \mathcal{S}^t_{n_s}  \}_{t=1}^{n_t}$
be a set of longitudinal tensors $\{ \mathcal{S}^t \}_{t=1}^{n_t}$, where each tensor $\mathcal{S}^t \in \mathbb{R}^{n_r \times n_r \times n_v}$ represents a brain multigraph with $n_v$ views and $n_r$ nodes (\textbf{Fig.}~\ref{fig1}--A). Particularly, each edge is a vector $\mathbf{e}_{i,j}^t \in \mathbb{R}^{n_v}$ that connects the node $i$ to $j$ storing connectivity weights across the $n_v$ views. 

\textbf{B) Recurrent multigraph integrator network.} Next, we give an overview of our ReMI-Net architecture and detail each of its fundamental operations and components.  \emph{\textbf{1) View normalization layer.}} Since the value range of connectivities in a brain multigraph might largely vary across views, the CBT learning task might be biased towards particular views. To address this issue, we first design a view-normalizer that normalizes the views between  0-1 for each training subject. Specifically, our normalizer includes a normalization layer from \cite{Ba:2016} that adds learnable parameters --weights and biases-- to the z-score normalization, then an additional sigmoid layer shifts the scale to 0-1. \emph{\textbf{2)  Cyclic recursion.}} Next, we input the \emph{normalized} multigraph at baseline timepoint $t_1$ to each of the time-dependent ReMI-Net cells, each learning the hidden-state embeddings $\mathbf{h}^t \in \mathbb{R}^{n_r \times n_h}$ of multigraph nodes across views at baseline and follow-up timepoints $t_i \geq t_1$. $n_h$ denotes the dimensionality of the embedding space. For initialization, we introduce a prior hidden-state vector $\mathbf{h}^0$ composed of zero elements for each node. To reinforce the learning of the baseline cell in our recurring architecture, we feed the hidden-state node embeddings from the last cell at $n_t$ back to the first one --noted as a single learning cycle.  \emph{\textbf{3) Subject-specific CBT prediction over time.}} After learning the hidden-state embeddings $\mathbf{h}^t$ at each timepoint $t$, we derive the subject-specific CBT $\mathbf{\hat{C}}^t_s$ for subject $s$ at timepoint $t$ by several tensor operations (\textbf{Fig.}~\ref{fig1}--B). First, we replicate $\mathbf{h}^t$ $n_r$ times producing a tensor  $\mathcal{R}^t \in \mathbb{R}^{n_r \times n_h \times n_r}$, which is then transposed along dimension $x$ and $z$ (i.e., transpose $\mathcal{R}_{xyz}^t$ to $\mathcal{R}_{zyx}^t$). Next, we compute the element-wise absolute difference between $\mathcal{R}^t$ and its transpose $\mathcal{R}^{t^T}$ and sum along the $y$ axis of the resulting tensor. This final matrix $\mathbf{\hat{C}}^t_s$ is the predicted subject-specific CBT at time $t$ from hidden-state node embedding matrix $\mathbf{h}^t$. Next, we  detail the inner working of each recurrent cell and how we derive the \emph{population-based} CBTs from the individual ones.

\textbf{C) Double RNN graph convolutional cell.} In each time-dependent cell, the hidden-state node embeddings $\mathbf{h}^t$ are propagated through a message passing network that performs a graph convolution operation for each node $i$, by calculating messages from each of its neighbors $j \in \mathcal{N}(i)$, where $\mathcal{N}(i)$ is the set of neighbors of node $i$. According to \cite{Gilmer:2017}, a message from node $j$ to $i$ is theoretically more effective if the message depends on both source and the destination (i.e., pair messages). Therefore, a pair message function can be defined in the spirit of \cite{Battaglia:2016}: $\mathbf{m}_{i,j}^{t+1} = \mathcal{M}(\mathbf{h}^t_i, \mathbf{h}^t_j, \mathbf{e}_{i,j}^t)$, where $\mathbf{m}_{i,j}^{t+1} \in \mathbb{R}^{n_h}$ is the message vector from $j \in \mathcal{N}(i)$ to node $i$ computed at the next timepoint. In order to compute the next hidden-state node embeddings in a graph convolutional fashion that captures nonlinear dependencies between nodes, we further propose a novel message function as: $\mathcal{M}(\mathbf{h}^t_i, \mathbf{h}^t_j, \mathbf{e}_{i,j}^t) = f_{\Theta}(\mathbf{e}_{i,j}^t,\mathbf{h}_{i}^t) \odot f_{\Theta}(\mathbf{e}_{i,j}^t, \mathbf{h}_{j}^t)$ (\textbf{Fig.}~\ref{fig1}--C). This is operated via each ReMI-Net cell composed of two RNN cells that share and pass messages to generate the hidden-state node embeddings at the next timepoint $t+1$. Both RNN cells input the baseline edge features $\mathbf{e}_{i,j}^{t=1}$ in addition to taking the hidden embeddings $\mathbf{h}^t_i$ for node $i$ and $\mathbf{h}^t_j$ for node $j$, respectively (\textbf{Fig.}~\ref{fig1}--C). Next, we perform an element-wise multiplication between the outputs of both RNN cells as a message to node $i$ from $j$. We then average the messages across neighbors to eventually generate the hidden-state embedding vector  $\mathbf{h}^{t+1}_i \in \mathbb{R}^{n_h}$  for node $i$ at follow-up timepoint:

\centerline{$\mathbf{h}_{i}^{t+1} = \Theta \cdot \mathbf{h}_{i}^t + \frac{1}{|\mathcal{N}(i)|} \left [\sum_{j \in \mathcal{N}(i)} f_{\Theta}(\mathbf{e}_{i,j}^t,\mathbf{h}_{i}^t) \odot f_{\Theta}(\mathbf{e}_{i,j}^t, \mathbf{h}_{j}^t) \right ]$;}

$f_\Theta$ is the proposed RNN cell function to learn, which maps for each node a pair of edge features and corresponding hidden state embeddings to an $n_h$-dimensional vector. A separate set of learnable parameters --weights and biases-- ($\Theta \in \mathbb{R}^{n_h \times n_h}$) is defined to generalize the convolution and consider the case of the node level effects \cite{Battaglia:2016}. Our designed $f_\Theta$ function is also generalizable to long-short term memory (LSTM) \cite{yu2019review} and gated recurrent unit (GRU) \cite{chung2014empirical} networks.

\textbf{D) Time-dependent loss.} To learn a centered and well-representative CBT, we randomly sample $k$ subjects from the input population and take the Frobenius distance ($d_F(A,B) = \sqrt{\sum_{i}\sum_{j}|A_{ij} - B_{ij}|^2}$ ) between each subject-specific CBT $\mathbf{\hat{C}}^t_s$ and each random training subject (\textbf{Fig.}~\ref{fig1}--D). Next, we sum the resulting distances across random subjects and average across timepoints. The random training sampling has a regularization effect as it reduces the risk of overfitting \cite{Gurbuz:2020}. We formulate our \emph{subject-specific} CBT centeredness loss for subject $s$ as follows: 

    \centerline{$\textcolor{violet}{\mathcal{L}_c(s)} = \frac{1}{n_t} \sum_{t=1}^{n_t} \sum_{v=1}^{n_v} \sum_{\mathcal{S}_k^t \in \mathbb{K}}^{n_k} \lambda_v || \mathbf{\hat{C}}^t_s - \mathcal{S}^t_k ||_F; \quad
    \lambda_{v} = \frac{max \left \{ \mu_{v} \right \}_{v=1}^{n_{v}}}{\mu_{v}}$}

where $\lambda_v$ is a view-normalizer calculated for each subject and each timepoint separately, $\mathbb{K}$ is a randomly selected subset and $\mathcal{S}_k^t$ is a random training sample. $\mu_v$ is the mean value of each view. To stabilize our training over time, we further propose a time regularization loss where we constraint two consecutive CBTs to be similar as we hypothesize that connectivity brain changes are local and sparse:

\centerline{$\textcolor{coral}{\mathcal{L}_t(s)} = \frac{1}{n_t - 1}\sum^{n_t - 1}_{t=1}\sqrt{\sum_{i}\sum_{j}(\mathbf{\hat{C}}^{t+1}_{s_{i,j}} - \mathbf{\hat{C}}^{t}_{s_{i,j}})^2}$}

Lastly, time regularizer loss $\mathcal{L}_t$ has also a weight $\alpha$ as a hyperparameter in total loss: $\mathcal{L}_{total} = \frac{1}{n_s}\sum^{n_s}_{s=1} [\textcolor{violet}{\mathcal{L}_c(s)} + \alpha \textcolor{coral}{\mathcal{L}_t(s)} ]$.

\textbf{E) Generation of the \emph{population-specific} CBT evolution trajectory.} So far, each of the estimated subject-specific CBTs is somewhat biased towards a specific subject. Hence, we perform an additional post-training operation by passing all training samples through the trained ReMI-Net to generate subject-specific CBTs (\textbf{Fig.}~\ref{fig1}--E). Since we aim to obtain population representative CBTs at all timepoints $t$, we use the element-wise median to select the most centered connectivities across all subject specific CBTs. This results in  population center $\mathbf{\hat{P}}^t$ at time $t$. 

\section{Results and Discussion}

\textbf{Longitudinal brain multigraph dataset.} We evaluated our architecture on 67 subjects (32 subjects diagnosed with Alzheimer's
Diseases (AD) and 35 with Late Mild Cognitive Impairment (LMCI)) from the Alzheimer's Disease Neuroimaging Initiative (ADNI) dataset \cite{ADdataset}, each represented by four cortical brain networks derived from T1-weighted magnetic resonance imaging for each cortical hemisphere, respectively. Each network comprises  $n_r = 35$ nodes and is measured at $n_t = 2$ timepoints (baseline and a 6-month follow up). The multigraph tensor views are derived from the following measures: maximum principal curvature, the mean cortical thickness, the mean sulcal depth, and the average curvature. Each edge weight encodes the average dissimilarity in morphology between two brain cortical ROIs using Desikan-Killiany cortical surface parcellation atlas. 

\textbf{Evaluation and parameter setting.} We evaluated our model against its variants and benchmark methods using 5-fold cross-validation where we learned the CBT from all samples but for each hemisphere independently. The recurrent block is composed of 3 layers of ReMI-Net cells, each with hidden-state node embeddings size  12, 36 and 24, respectively. All models are trained using gradient descent with Adam optimization, a learning rate of 0.0008 and 100 epochs. We empirically set the hyperparameter $\alpha = 0.3$ in our loss function and the number of random training samples in the centeredness loss to $k=10$. As for the recursive cycling, we only cycle once. 

\textbf{Benchmarks and ReMI-Net variations.} Since our ReMI-Net is the first method aiming to forecast CBT evolution from a single baseline population, we benchmarked our model against DGN \cite{Gurbuz:2020} at each timepoint independently. We further evaluated 3 variants of our model: \textbf{(1) Standard vanilla:} 3 layers of graph convolutional cells without normalization and cyclic recursion to the baseline timepoint; \textbf{(2) Standard cyclic:} 3 layers of graph convolutional cells with cyclic recursion operation and without normalization; and \textbf{(3) Cyclic min-max norm:} 3 layers of graph convolutional cells with cyclic recursion and min-max view normalization. 

\begin{table}
	\centering
	\caption{\emph{Reproducibility rate between ROIs selected by CBT-based methods and an independent learner \cite{Varma:2009}.}}
	\begin{tabular}{c c c c}
		\hline\noalign{\smallskip}
		Overlap Rate & netNorm \cite{Dhifallah:2020}  & DGN \cite{Gurbuz:2020} & \bf{ReMI-Net} \\
		\hline\noalign{\smallskip}
		AD-LMCI Left Hem. & 0.60 & \bf{0.73} & \bf{0.73} \\
		AD-LMCI Right Hem. & 0.33 & 0.40 & \bf{0.80} \\
		\hline\noalign{\smallskip}
	\end{tabular}
\label{tab:1}
\end{table}

\begin{figure}[ht!]
\centering
\includegraphics[width=12.5cm]{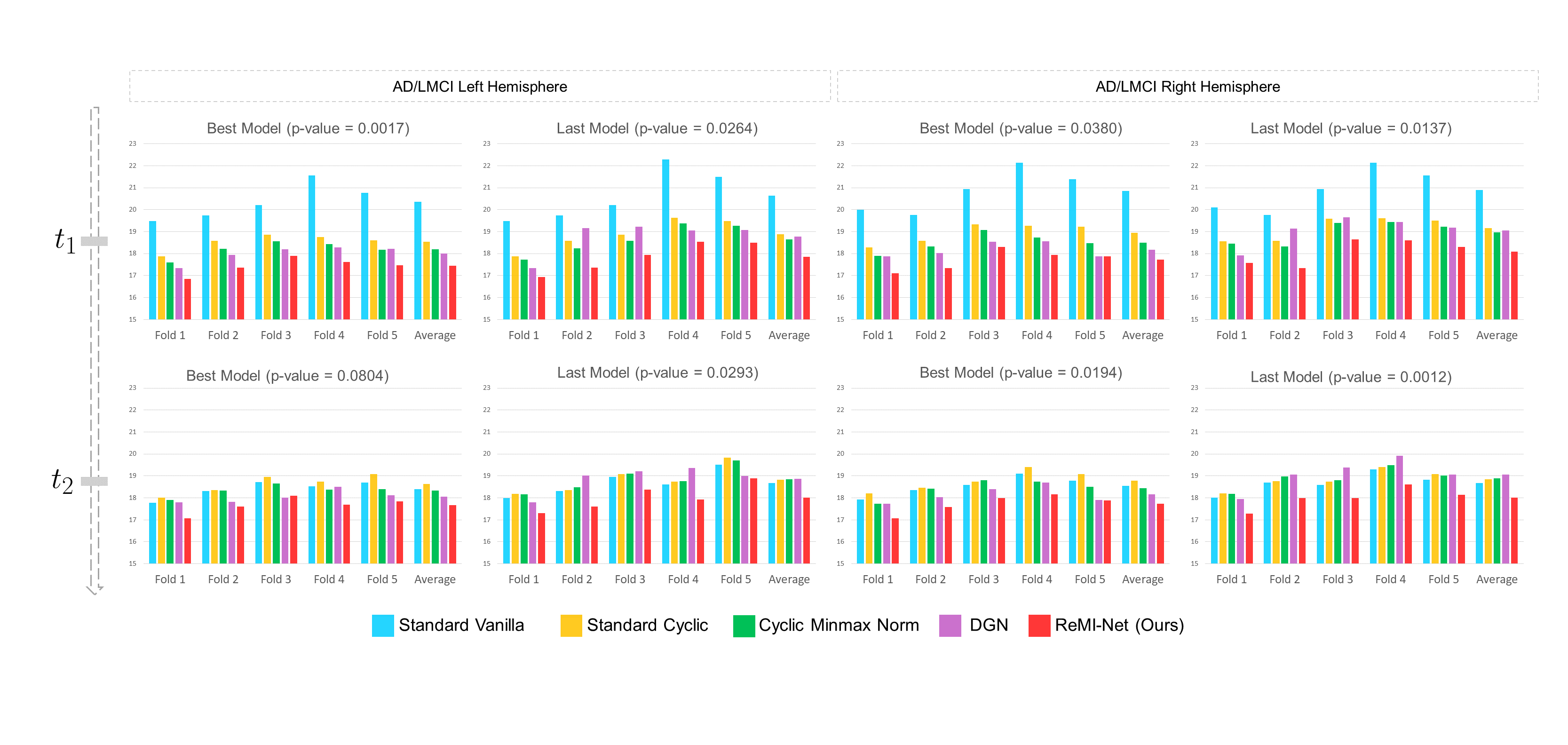}
\caption{\emph{Evaluation of the centeredness of the learned CBTs by DGN \cite{Gurbuz:2020} and ReMI-Net variants at baseline and follow-up timepoints $t_1$ and $t_2$, respectively.}}
\label{fig:2}
\end{figure}

\begin{figure}
    \centering
    \includegraphics[width=12cm]{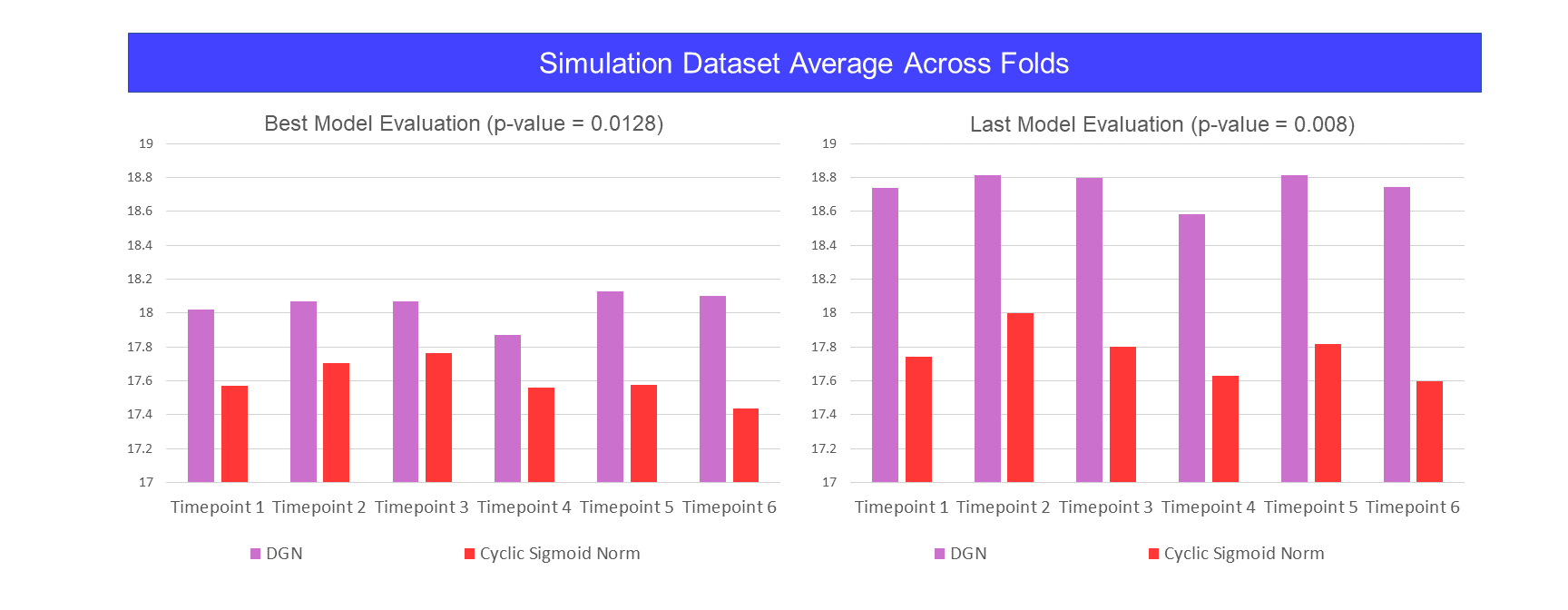}
    \caption{This chart displays the centeredness results on the simulation dataset for each timepoint averaged across folds. }
    \label{fig_sim}
\end{figure}

\textbf{CBT representativeness test.} To evaluate our ReMI-Net at different timepoints, we calculated the average Frobenius distance between the predicted training CBT (i.e., generated from the training set) and all samples in the testing set. We inspect the results at each timepoint independently using two different evaluation strategies. \textbf{(1) Last model:} We train each model for a complete 100 epochs for easy comparison across methods and \textbf{(2) Best model:}  We pick the best model that achieves the best results on the testing samples by adopting an early stop training strategy (i.e., number of epochs $<$ 100). \textbf{Fig.}~\ref{fig:2} displays the average Frobenius distance quantifying the centeredness and representativeness of the CBT forecasted by ReMI-Net and its 3 variants at each timepoint. \textbf{\textbf{Table.}~\ref{tab_mae} 1-supp} displays the MAE in node strength where our ReMI-Net outperforms benchmarks.  \textbf{Fig.}~\ref{fig:fig_sim} also shows that our model achieves the best performance on a simulated connectomic dataset with 6 timepoints. We also report the CBT results by training DGN \cite{Gurbuz:2020} at each timepoint, \emph{independently}. Lastly, we report the p-values using a two-tailed paired t-test between DGN (the best performing method among benchmarks) and our ReMI-Net. Clearly, our  model significantly outperforms DGN and its variants in terms of CBT representativeness across all test folds, evaluation datasets (left and right hemispheres) and at all timepoints.

\begin{table}[]
\centering
    \caption{Mean absolute errors between the node strengths of predicted CBTs and the dataset.}
    \begin{tabular}{ccc}
        \hline\noalign{\smallskip}
        Left Hemisphere         & AD                & LMCI          \\ \hline\noalign{\smallskip}
        DGN \cite{Gurbuz:2020}  & 3.12              & 2.48          \\
        \textbf{ReMI-Net}       & \textbf{1.33}     & \textbf{1.32} \\ \hline\noalign{\smallskip}
    \end{tabular}
    \begin{tabular}{ccc}
        \hline\noalign{\smallskip}
        Right Hemisphere        & AD                & LMCI          \\ \hline\noalign{\smallskip}
        DGN \cite{Gurbuz:2020}   & 3.21              & 2.37          \\
        \textbf{ReMI-Net}       &  \textbf{1.40}    & \textbf{1.47} \\ \hline\noalign{\smallskip}
    \end{tabular}
    \label{tab_mae}
\end{table}

\begin{table}
\centering
\caption{The most discriminative ROIs (top to bottom) between AD and LMCI patients for left and right hemispheres at baseline and follow-up timepoints.}
\scalebox{0.75}{
    \begin{tabular}{cccc}
        \multicolumn{2}{c}{\textbf{Timepoint $t_1$}}           & \multicolumn{2}{c}{\textbf{Timepoint $t_2$}}                 \\\hline \noalign{\smallskip}
        \textbf{Left Hemisphere} & \textbf{Right Hemisphere}& \textbf{Left Hemisphere}  & \textbf{Right Hemisphere}     \\\hline \noalign{\smallskip}
        \cellcolor{pink!40}{Entorhinal cortex} & \cellcolor{pink!40}{Entorhinal cortex} & \cellcolor{pink!40}{Entorhinal cortex} & \cellcolor{pink!40}{Entorhinal cortex}                 \\
        \cellcolor{orange!30}{Inferior temporal gyrus}             & \cellcolor{yellow!40}{Pericalcarine cortex} & \cellcolor{orange!30}{Inferior temporal gyrus} & \cellcolor{yellow!40}{Pericalcarine cortex}              \\
        \cellcolor{green!30}{Transverse temporal cortex}          & \cellcolor{green!30}{Transverse temporal cortex}        & \cellcolor{red!30}{Rostral anterior cingulate cortex} & \cellcolor{green!30}{Transverse temporal cortex}        \\
        \cellcolor{red!30}{Rostral anterior cingulate cortex}   & \cellcolor{red!30}{Rostral anterior cingulate cortex} & \cellcolor{green!30}{Transverse temporal cortex}        & \cellcolor{red!30}{Rostral anterior cingulate cortex} \\
        \cellcolor{blue!20}{Caudal anterior cingulate cortex}    & \cellcolor{orange!30}{Inferior temporal gyrus}           & \cellcolor{blue!20}{Caudal anterior cingulate cortex}  & \cellcolor{purple!30}{Insula cortex}
        \\\noalign{\smallskip}\hline \hline
        
    \end{tabular}{\smallskip}
}
\label{tab:2}
\end{table}

\begin{table}
\centering
\caption{The top brain ROIs marking the difference between the baseline and follow-up CBTs generated for AD patients by our ReMI-NET.}
\scalebox{0.9}{
    \begin{tabular}{cc}
    \hline\noalign{\smallskip}
    
        \textbf{Left Hemisphere}                    & \textbf{Right Hemisphere}                             \\\hline \noalign{\smallskip}
        \cellcolor{yellow!40}{Pericalcarine cortex}                        & \cellcolor{yellow!40}{Pericalcarine cortex}                                  \\
        \cellcolor{blue!20}{Caudal anterior cingulate cortex}            & \cellcolor{blue!20}{Caudal anterior cingulate cortex}                      \\
        \cellcolor{cyan!20}{Medial orbital frontal cortex}               & \cellcolor{cyan!20}{Medial orbital frontal cortex}                         \\
        \cellcolor{gray!20}{Postcentral gyrus}                           & \cellcolor{gray!20}{Postcentral gyrus}                                     \\
        \cellcolor{blue!30}{Precuneus cortex}                            & \cellcolor{blue!30}{Precuneus cortex}                                      \\\noalign{\smallskip} \hline
    \end{tabular}{\smallskip}
} 
\label{tab:3}
\end{table}

\textbf{CBT discriminativeness and biomarker reproducibility test.} By acting as a connectional brain fingerprint, a well-centered CBT can capture the most discriminative ROIs of a multigraph population between two brain states (e.g., LMCI \emph{vs.} AD) and identify their alternations over time caused by a progressive disease such as dementia. To test this hypothesis, we train our ReMI-Net on each class-specific population --LMCI and AD independently, to predict the CBT at baseline and follow-up timepoints. Next, we compute the residual matrix between the CBT$^{LMCI}$ and CBT$^{AD}$ via absolute element-wise difference. Then we assign a discriminability score for each brain ROI by summing the elements of its row in the residual matrix. The ROIs with the highest discriminability scores better disentangle both brain states. To further evaluate the reproducibility of the discovered biomarker regions, we train an independent learner, specifically the generalized multiple kernel learning (GMKL)\cite{Varma:2009}  which classifies samples and learns a weight for each ROI using 5-fold cross-validation. Next, we inspect the reproducibility overlap rate of the top $K=15$ most discriminative ROIs between GMKL and each of CBT learning models. \textbf{Table.}~\ref{tab_mae}~\ref{tab:1} shows that ReMI-Net has a significant increase in the overlap reproducibility in the right hemisphere and a similar performance to state-of-the-art DGN \cite{Gurbuz:2020} in the left hemisphere.

\textbf{Clinical discoveries.} \textbf{1) \emph{Discriminability approach}.} \textbf{Table.}~\ref{tab_mae}~\ref{tab:2} displays the top 5 most discriminative between AD and LMCI at baseline and follow-up timepoints $t_1$ and $t_2$. The entorhinal cortex is found as the most discriminative ROI between both groups, which replicates the findings of independent clinical studies on LMCI and AD \cite{Yang:2019,Zhou:2016,Howett:2019}. \textbf{2) \emph{Progressive approach}.} \textbf{Table.}~\ref{tab_mae}~\ref{tab:3} shows the most affected ROIs by AD with the disorder progression over time. The results are significantly similar between left and right hemispheres. As reported in \cite{Yang:2019}, we also identify the pericalcarine cortex, precuneus cortex and medial orbital frontal cortex as the most dynamically changing regions in AD patients.

\section{Conclusion}

In this paper, we introduced the \emph{first} Recurrent Multi-graph Integrator Network to forecast the connectional brain template evolution over time using a single baseline population. The proposed method outperformed both state-of-the-art and variant benchmarks in terms of all evaluation measures. Moreover, we introduced our own recurrent graph convolution method and demonstrated the high reproducibility of biomarkers revealed by our learned time-dependent CBTs in a demented population. In our future work, we will evaluate the generalizability of ReMI-Net to large-scale brain multigraph populations derived from different neuroimaging modalities. We will also explore other operations to derive the CBT from the learned embeddings including graph-based similarity measures.

\section{Acknowledgments}

I. Rekik is supported by the European Union's Horizon 2020 research and innovation programme under the Marie Sklodowska-Curie Individual Fellowship grant agreement No 101003403 (\url{http://basira-lab.com/normnets/}) and the Scientific and Technological Research Council of Turkey under the TUBITAK 2232 Fellowship for Outstanding Researchers (no. 118C288, \url{http://basira-lab.com/reprime/}). However, all scientific contributions made in this project are owned and approved solely by the authors.

\section{Supplementary material}

We provide three supplementary items for reproducible and open science:

\begin{enumerate}
	\item A 5-mn YouTube video explaining how our framework works on BASIRA YouTube channel at \url{https://youtu.be/tthw51zlxXo}.
	\item ReMI-Net code in Python on GitHub at \url{https://github.com/basiralab/ReMI-Net}. 
\end{enumerate}

\bibliographystyle{splncs}
\bibliography{Biblio}

\end{document}